\documentclass[11pt]{article}
\usepackage{acl2016}
\usepackage{times}
\usepackage{url}
\usepackage{latexsym}
\usepackage{amsmath}
\usepackage{amsfonts}
\usepackage{amssymb}
\usepackage{graphicx}
\usepackage[caption=false,font=footnotesize]{subfig}
\usepackage{cleveref}
\usepackage[T1]{fontenc}

\newcommand{\ignore}[1]{}

\aclfinalcopy % Uncomment this line for the final submission
\usepackage{url}
\usepackage{dsttr}
\begin{document}

%\title{  Interactive Learning through Dialogue for Language Grounding using  zero-shot learning  }

\title{Training an adaptive dialogue policy for interactive learning of visually grounded word meanings}
 

\author{Yanchao Yu\\
  Interaction Lab   \\
 Heriot-Watt University  \\
 % Affiliation / Address line 3 \\
  {\tt y.yu@hw.ac.uk} \\ \And
  Arash Eshghi \\
 Interaction Lab   \\
 Heriot-Watt University  \\
 % Affiliation / Address line 3 \\
  {\tt a.eshghi@hw.ac.uk} \\ \And
  Oliver Lemon \\
 Interaction Lab   \\
 Heriot-Watt University  \\
 % Affiliation / Address line 3 \\
  {\tt o.lemon@hw.ac.uk}}

 %\author{Anonymous}

% OL comments / edits appear tagged as "%OL"
\maketitle

\begin{abstract}

We present a multi-modal dialogue system for interactive learning of perceptually grounded word meanings from a human tutor. The system integrates an incremental, semantic parsing/generation framework - Dynamic Syntax and Type Theory with Records (DS-TTR) - with a set of visual classifiers that are learned throughout the interaction and which ground the meaning representations that it produces. We use this system in interaction with a simulated human tutor to study the effects of different dialogue policies and capabilities on accuracy of learned meanings, learning rates, and efforts/costs to the tutor. We show that  the overall performance of the learning agent is affected by (1) who takes initiative in the dialogues; (2) the ability to express/use their confidence level about   visual attributes; and (3) the ability to process elliptical and incrementally constructed dialogue turns.  Ultimately, we train an adaptive dialogue policy which optimises the trade-off between classifier accuracy and tutoring costs.

%Human tutors can correct, question, and confirm the statements of a dialogue agent which is trying to interactively learn the meanings of perceptual words, e.g.\ colours and shapes.
%We show that different learner and tutor dialogue strategies lead to different learning rates, accuracy of learned meanings, and effort/costs for   human tutors. For example, we show that a learner which can handle corrections in dialogue can learn meanings that are as accurate as a fully-supervised learner, but with less cost/effort to the human tutor.

%multi-attribute objects through Human-Robot Interaction. 
%We design a semantic and visual processing system to support this and illustrate how they can be integrated. 
%, where we assume that the system does not know the meanings of words
% from human descriptions
% before learning. 
%Previous work could only learn about one attribute in each dialogue turn, whereas here we allow multiple labels to be learned in each turn (e.g.\ ``This is a blue book"). 
%An attribute-based label set is built using Type Theory with Records (TTR) semantic representations and will then be used to interactively train image classifiers (e.g.\ ``this is a red mug").
% (MkLNN and TRAM) and a `zero-shot' learning method with visual features.
 %We then evaluate the effectiveness of dialogue interaction on the task of attribute-based object and word-meaning learning. We also test 

\end{abstract}

\section{Introduction}\label{tab:introduction}
Identifying, classifying, and talking about objects or events in the surrounding environment are key capabilities for intelligent, goal-driven systems that interact with other agents and the external world (e.g.\  robots,  smart spaces, and other automated systems). To this end, there has recently been a surge of interest and significant progress made on a variety of related tasks, including generation of Natural Language (NL) descriptions of images, or identifying images based on NL descriptions  \cite{karpathy2014deep,bruni2014multimodal,socher2014grounded,Farhadi09describingobjects,silberer-lapata:2014:P14-1,sun2013attribute}. %Another strand of work has focused on learning to generate object descriptions and object classification based on low level concepts/features (such as colour, shape and material), enabling systems to identify and describe novel, unseen images \cite{Farhadi09describingobjects,silberer-lapata:2014:P14-1,sun2013attribute}.

\begin{figure}\centering\begin{small}\hspace{-0.5cm}
% \begin{tabular}{|p{3cm}|c|l|}\hline
% \textbf{Dialogue}& \textbf{Image} & \textbf{Final Semantics in TTR}\\
% \raisebox{0.3cm}{\begin{tabular}{l}
% T(utor): What is this?\\
% L(earner): a red circle?\\
% T: no, a red square.\\
% L: oh, okay.

% \end{tabular}}
% & \raisebox{-0.5cm}{\includegraphics[width=1.8cm]{imageResource/red_square.jpg}} & \raisebox{0.3cm}{$\ttrnode{}{x_{=o1}&e\\p2&red(x)\\p3&square(x)}$}\\
% &&\\

% \hline
% &&\\
% \begin{tabular}{l}

% T: What can you see?\\
% L: something orange.\\
% T: What shape is it?\\
% L: a square.\\
% T: no, it's a circle.\\
% L: uhu

% \end{tabular}

% & \raisebox{-0.5cm}{\includegraphics[width=1.8cm]{imageResource/orange_circle.jpg}} & $\ttrnode{}{x1_{=o2}&e\\s_{=S}&per\\p&circle(x1)\\p1&orange(x1)\\p2&see(s,x1)}$\\\hline
% \end{tabular}
\includegraphics[width=\linewidth]{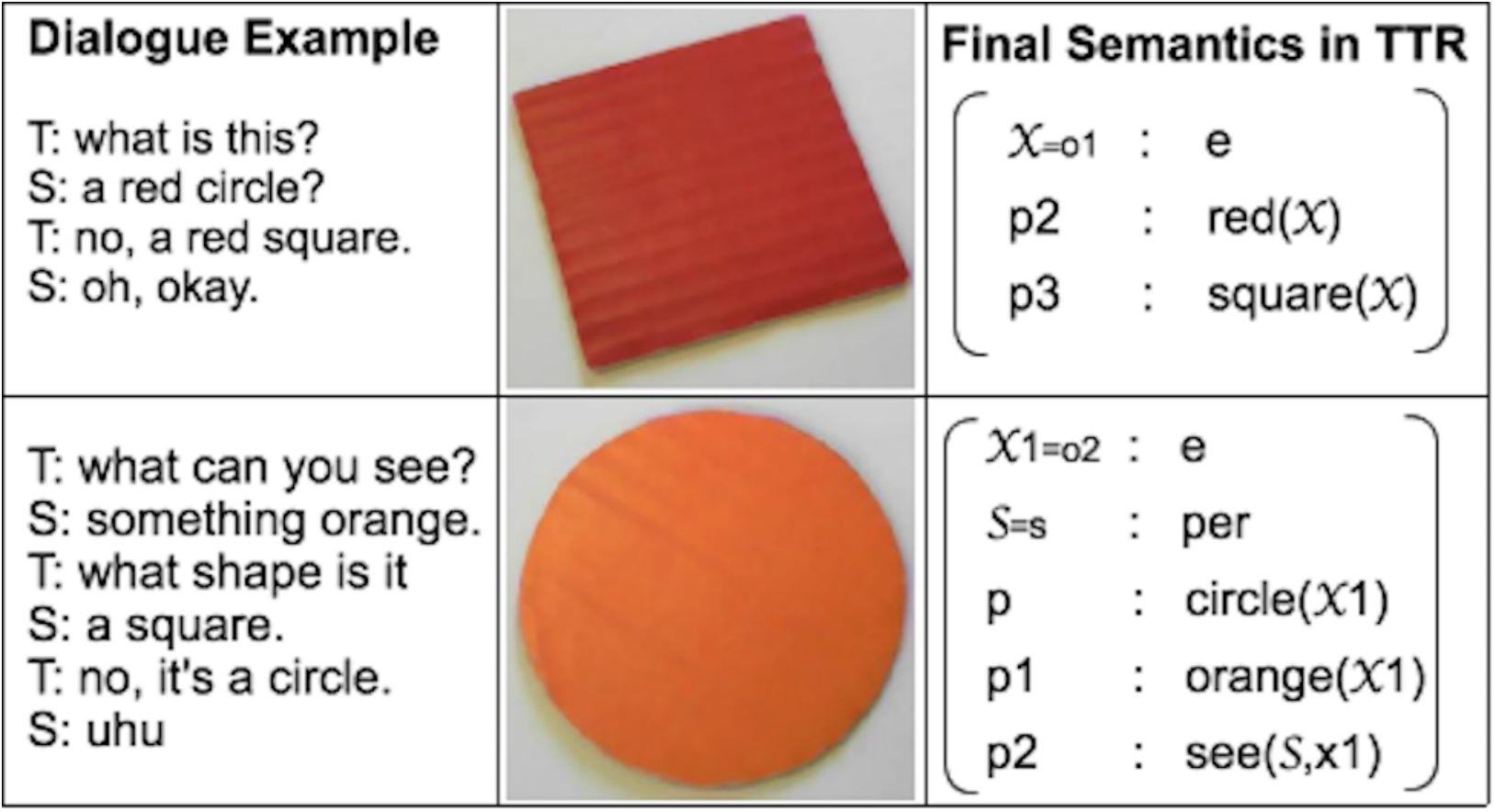}
\caption{Example dialogues \& interactively agreed semantic contents.}
\vspace{-0.7cm}
\label{fig:example}\end{small}
\end{figure}

Our goal is to build  {\it interactive} systems that can learn grounded word meanings relating to their perceptions of real-world objects -- this is different from previous work such as e.g.\ \cite{roy2002describer}, that learn groundings from descriptions without any interaction, and more recent work using Deep Learning methods (e.g.\ \cite{socher2014grounded}). 

Most of these systems rely on training data of high quantity with no possibility of online error correction. Furthermore, they are unsuitable for robots and multimodal systems that need to continuously, and incrementally learn from the environment, and may encounter objects they haven't seen in training data. These limitations are likely to be alleviated if systems can learn concepts, as and when needed, from situated dialogue with humans. Interaction with a human tutor also enables systems to take initiative and seek the particular information they need or lack by e.g.\ asking questions with the highest information gain (see e.g.\ \cite{DBLP:conf/iros/SkocajKVMJKHHKZZ11}, and Fig.~\ref{fig:example}). For example, a robot could ask questions to learn the colour of a ``square" or to request to be presented with more ``red" things to improve its performance on the concept (see e.g.\ Fig.\ ~\ref{fig:example}). Furthermore, such systems could allow for meaning negotiation in the form of clarification interactions with the tutor.

This setting means that the system must be {\it trainable from little data, compositional, adaptive, and able to handle natural human dialogue with all its glorious context-sensitivity and messiness} -- for instance so that it can  learn visual concepts suitable for specific tasks/domains, or even those specific to a particular user.
%OL added the sentence below to link with later text
Interactive systems that learn continuously, and over the long run from humans need to do so \textit{incrementally}, \textit{ quickly}, and \textit{with minimal effort/cost to  human tutors}. 

\begin{figure*}[!t]
\vspace{-0.2cm}
	\centering
	\includegraphics[width=.75\linewidth]{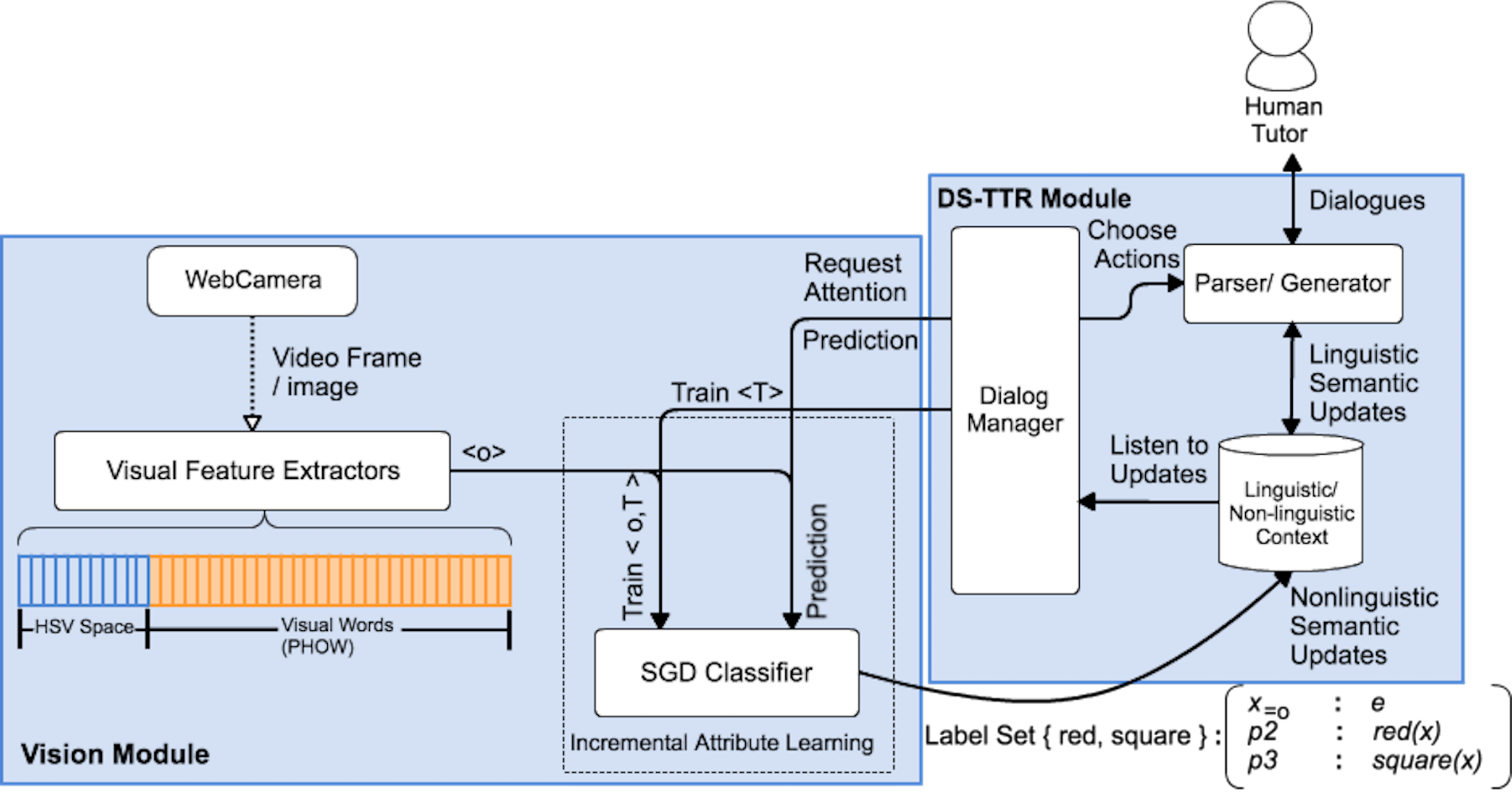}
	\vspace{-0.2cm}
    \caption{Architecture of the teachable system}\label{fig:architecture}
\end{figure*}

In this paper, we first outline an implemented dialogue system that integrates an incremental, semantic grammar framework, especially suited to dialogue processing -- Dynamic Syntax and Type Theory with Records (DS-TTR\footnote{Download from \url{http://dylan.sourceforge.net}} \cite{Kempson.etal01,Eshghi.etal12}) with visual classifiers which are learned during the interaction, and which  provide perceptual grounding for the basic semantic atoms in the semantic representations (Record Types in TTR) produced by the parser (see Fig.~\ref{fig:example}, Fig.~\ref{fig:architecture} and section~\ref{sec:architecture}). 
%AE: commented out the below. 
%In effect, the dialogue with the tutor continuously provides semantic information about objects in the scene which is then fed to online classifiers in the form of training instances. Conversely, the system can utilise the grammar and its existing knowledge about the world, encoded in the meanings it has already learned, to make reference to and formulate questions about the different attributes of an object identified in the visual scene.\footnote{Here we assume that the words being grounded are in the lexicon, i.e.\ that their syntactic and semantic type are known: we leave the problem of grammar induction to one side here, though see \cite{Eshghi.etal13CMCL}}.

We then   use this system in interaction with a simulated human tutor, to test hypotheses about how the accuracy of learned meanings, learning rates, and the overall cost/effort for the human tutor are affected by different dialogue policies and capabilities: (1) who takes {\bf initiative} in the dialogues; (2) the agent's ability to utilise their level of {\bf uncertainty} about an object's attributes; and (3) their ability to process {\bf elliptical as well as incrementally constructed dialogue turns}. The results show that differences along these dimensions have significant impact both on the accuracy of the grounded word meanings that are learned, and the processing effort required  by the tutors.

In section \ref{sec:adaptive} we train an adaptive dialogue strategy that finds a better trade-off between classifier accuracy and tutor cost.

\section{Related work} \label{sec:related_work}
In this section, we will present an overview of vision and language processing systems, as well as multi-modal systems that learn to associate them. We compare them along two main dimensions: \textit{Visual Classification methods: offline vs.\ online} and \textit{the kinds of representation learned/used}. %In this paper, we are more concerned with learning low-level attributes (e.g. colour and shape) than object categories. 

\vspace{-0.2cm}
\paragraph{Online vs.\ Offline Learning.} A number of implemented systems have shown good performance on classification as well as NL-description of novel physical objects and their attributes, either using offline methods as in ~\cite{Farhadi09describingobjects,DBLP:journals/pami/LampertNH14,DBLP:conf/nips/SocherGMN13,DBLP:journals/tkde/KongNZ13}, or through an incremental learning process, where the system's parameters are updated after each training example is presented to the system ~\cite{DBLP:journals/nn/FuraoH06,DBLP:journals/nca/ZhengSFZ13,DBLP:journals/tcyb/KristanL14}. For the interactive learning task presented here, only the latter is appropriate, as the system is expected to learn from its interactions with a human tutor over a period of time. Shen \& Hasegawa \shortcite{DBLP:journals/nn/FuraoH06} propose the SOINN-SVM model that re-trains linear SVM classifiers with data points that are clustered together with all the examples seen so far. The clustering is done incrementally, but the system needs to keep all the examples so far in memory. Kristian \& Leonardis \shortcite{DBLP:journals/tcyb/KristanL14}, on the other hand, propose the oKDE model that continuously learns categorical knowledge about visual attributes as probability distributions over the categories (e.g.\ colours). However, when learning from scratch, it is unrealistic to predefine these concept groups (e.g.\ that red, blue, and green are colours). Systems need to learn for themselves that, e.g.\ colour is grounded in a specific sub-space of an object's features. For the visual classifiers, we  therefore assume no such category groupings here, and instead learn individual binary classifiers for each visual attribute (see section \ref{sub:classification} for details). 

%AE: this is too much detail for the related work section 
%simple but efficient method that incrementally learns linear classifiers under convex loss functions (e.g. logistic regression) -- Stochastic Gradient Decent (SGD) classifier ~\cite{zhang2004solving} (see Section \ref{sub:classification}). 

\parskip 0pt
\paragraph{Distributional vs.\ Logical Representations.} Learning to ground natural language in perception is one of the fundamental problems in Artificial Intelligence. There are two main strands of work that address this problem: (1) those that learn distributional representations using Deep Learning methods: this often works by projecting vector representations from different modalities (e.g.\ vision and language) into the same space in order to be able to retrieve one from the other
%with loss functions designed to minimise the distance between them, thus allowing generation from one modality to the other 
~\cite{socher2014grounded,DBLP:conf/cvpr/KarpathyL15,silberer-lapata:2014:P14-1}; (2) those that attempt to ground symbolic logical forms, obtained through semantic parsing ~\cite{DBLP:journals/ml/TellexTJR14,kollar2013toward,DBLP:conf/aaai/MatuszekBZF14} in classifiers of various entities types/events/relations in a segment of an image or a video. Perhaps one advantage of the latter over the former method, is that it is strictly compositional, i.e.\ the contribution of the meaning of an individual word, or semantic atom, to the whole representation is clear, whereas this is hard to say about the distributional models. As noted, our work also uses the latter methodology, though it is dialogue, rather than sentence semantics that we care about.
Most similar to our work is probably that of  Kennington \& Schlangen \shortcite{Kennington.Schlangen15} who learn a mapping between individual words - rather than logical atoms - and low-level visual features (e.g.\ colour-values) directly. The system is compositional, yet does not use a grammar (the compositions are defined by hand). Further, the groundings are learned from pairings of object references in NL and images rather than from dialogue. 

What sets our approach apart from others is: a) that we use a domain-general, incremental semantic grammar with principled mechanisms for parsing and generation; b) Given the DS model of dialogue~\cite{Eshghi.etal15}, representations are constructed jointly and interactively by the tutor and system over the course of several turns (see Fig.~\ref{fig:example}); c) perception and NL-semantics are modelled in a single logical formalism (TTR); d) we effectively induce an ontology of atomic types in TTR, which can be combined in arbitrarily complex ways for generation of complex descriptions of arbitrarily complex visual scenes (see e.g.\ \cite{Dobnik.etal12} and compare this with \cite{Kennington.Schlangen15}, who do not use a grammar and therefore do not have logical structure over grounded meanings).
\ignore{%arash: replacing with our new related work section
%\vspace{-0.2cm}
\section{Related work} \label{sec:related_work}

% Compositional Semantic Grounding Techniques
There has been much recent research into learning to classify and describe images/objects. Some approaches attempt to ground meaning of words/phrases/sentences in images/objects by mapping these modalities into the same vector space \cite{karpathy2014deep,silberer-lapata:2014:P14-1,kiros2014multimodal}, or using distributional semantic models that build distributional representations with the conjunction of textual and visual information \cite{bruni2014multimodal}. 
%Other approaches, such as \cite{socher2014grounded}, propose Neural Network models based on Dependency Trees (DT), which project all words in a sentence into a DT structured representation to explore parents of each node and correlations between nodes.

In contrast to such approaches, which do not support NL dialogues, some approaches are designed based on logical semantic representations and some of them are incorporated with spoken dialogue systems \cite{DBLP:conf/iros/SkocajKVMJKHHKZZ11,UW_RSE_ICML2012,kollar2013toward}.  However, none of these approaches explores different dialogue strategies and their effects on learning, and on tutor effort, as we do in this paper.

%A well-known logical semantic parser is the Combinatory Categorial Grammar (CCG) parser, which represents natural language sentences from human tutors in the logical forms. The ``Logical Semantics with Perception" (LSP) framework by Kollar et al.\ \cite{DBLP:journals/tacl/KrishnamurthyK13} and the joint language/perception model by Matuszek et al.\ \cite{UW_RSE_ICML2012} are based on a CCG parser or using a CCG lexicon respectively. Although a CCG parser could generate similar logical representations to the DS-TTR parser/generator we use here, we believe that DS-TTR would show better performance than CCG in terms of handling the inherent incremental, fragmentary and highly context-dependent nature of dialogue.

%I've commented out the 'describer' system. Isn't it too old? 
%The ``Describer'' system \cite{roy2002describer} learns to generate image descriptions, but it works at the level of word sequences rather than logical semantics, and uses only synthetically generated scenes rather than real images and image processing. 

Our approach extends \cite{Dobnik.etal12} in integrating perception - vision in this case - and language within a single formal system: Type Theory with Records (TTR \cite{Cooper05}). The combination of deep semantic representations in TTR with an incremental grammar (Dynamic Syntax, \cite{Kempson.etal01}) allows complex multi-turn dialogues - including natural correction (e.g.\ ``A: Is this a square? B: No, a triangle.'') and clarification subdialogues - to be processed for interactive Natural Language grounding.
%, in contrast to systems such as \cite{DBLP:conf/iros/SkocajKVMJKHHKZZ11} which do not support incremental parsing. 

%OL - I moved this - it was originally in 3.3. Integration
What sets our approach apart from other work is: (1) that we use a domain-general, semantic grammar, with principled mechanisms for parsing and generation; (2) Given the DS model of feedback in dialogue \cite{Eshghi.etal15}, representations are constructed jointly and interactively so that both the system and the tutor can contribute to a single representation over the course of several turns (see e.g.\ second row of Fig.~\ref{fig:example}); (3) perception (vision) and NL-semantics are modelled in a single formalism (TTR), which affords inference as well as all the power that is afforded by all logical formalisms; (4) We effectively induce an ontology of atomic types in TTR, which can be combined in arbitrarily complex ways for generation of arbitrarily complex descriptions.

\subsection{Attribute classification}
  Regarding attribute-based classification or description, Farhadi et al.\ \shortcite{Farhadi09describingobjects} have successfully described  objects with attributes by sharing appearance attributes across object categories. Silberer and Lapata\ \shortcite{silberer-lapata:2014:P14-1} extend Farhadi et al.'s work to predict attributes using L2-loss linear SVMs and to learn the associations between visual attributes and particular words using Auto-encoders. Sun et al.\ \shortcite{sun2013attribute} also build an attribute-based identification model based on hierarchical sparse coding with a K-SVD algorithm, which recognizes each attribute type using multinomial logistic regression. However, as these models require a large mass of training data, an increasing amount of research attempts to learn novel objects using `one-shot' \cite{DBLP:journals/pami/Fei-FeiFP06,DBLP:conf/aaai/KrauseZWS14} or `zero-shot' learning algorithms \cite{DBLP:journals/cviu/Fei-FeiFP07,DBLP:journals/pami/LampertNH14}. They enable a system to classify unseen objects with fewer or no examples by sharing {\it attributes} between known and unknown objects. 
%OLnew --> 
Note that these methods ultimately focus on object class labels, using attributes as intermediate representations.

On the other hand, to learn attribute-based objects through NL interaction, some approaches learn unknown objects or attributes with online incremental learning algorithms \cite{DBLP:journals/cviu/Fei-FeiFP07,DBLP:conf/cvpr/KankuekulKTH12}. 
%OL - we mention these papers bot don't say how our work is better thahn them!
The `George' system \cite{DBLP:conf/iros/SkocajKVMJKHHKZZ11}, which is similar in spirit to our work, learns  object attributes from a human tutor and creates   questions to request information to fill its knowledge gaps. %However, the George system %can only learn about one attribute in each turn and it
 %only learns about 2 shapes and 8 colours. Our goal is to couple attribute classifiers with much wider coverage to the  formal semantics of a full Natural Language dialogue system. 
However, this work does little in the way of dialogue processing, and only uses a single dialogue strategy for learning. 
}
%----------end of the replaced related work section

\section{System Architecture} 
\label{sec:architecture}
We have developed a system to support an attribute-based object learning process through natural, incremental spoken dialogue interaction. The architecture of the system is shown in Fig.~\ref{fig:architecture}. The system has two main modules: a vision module for visual feature extraction,  classification, and learning; and a dialogue system module using DS-TTR. %Visual feature representations are built based on base features akin to \cite{Farhadi09describingobjects}.
% We do not yet have a fully integrated dialogue system, so for our experiments presented below, we assume access to logical semantic representations, that will be output by the DS-TTR parser/generator as a result of processing dialogues with a human tutor (more on this below) -- and interface these representations with   attribute-based  image classifiers.
 Below we describe these components individually and then explain how they interact.
 
\subsection{Attribute-based Classifiers used} \label{sub:classification}

Yu et.\ al  \shortcite{yu-eshghi-lemon:2015:VL,yu-eshghi-lemon:2015:semdial} point out that neither multi-label classification models nor `zero-shot' learning models show acceptable performance on attribute-based learning tasks. Here, we instead use Logistic Regression SVM classifiers with Stochastic Gradient Descent (SGD) \cite{zhang2004solving} to incrementally learn attribute predictions. %\footnote{Although SGD models mainly execute a batch learning process that requires building classifier structures based on the whole dataset before iterative learning, the WEKA tool supports implementing an incremental version of SGD, which initializes model weights randomly and updates the weights immediately once a single unseen training sample presents.}. %

%AE: Comment out the footnote entirely? Too much detail?

%AE: I don't understand the following paragraph.... what do you mean by normalising probability distribution..... ??

%

All classifiers will output attribute-based label sets and corresponding probabilities for novel unseen images by predicting binary label vectors. 
\ignore{%OL 23/4/16 commented this out, as it will confuse the reader
As SGD classifier, as binary classifier, may produce the probability score independently for each attribute label, we need to manually normalize the probability distribution for each attribute category (i.e.\ colour and shape) separately after the original classification process, which may generate correlations between attribute classifiers with same category. (Note: a normalization constant is applied to ensure that the sum of prediction probabilities of all colour/shape attributes totals one).} % end ignore
 We build visual feature representations  to learn classifiers for particular attributes, as explained in the following subsections.

% All models will output attribute-based label sets for novel unseen images by predicting binary label vectors. We build visual representations and binary label vectors as inputs to train new classifiers for learning attributes, as explained in the following subsections. 

\subsubsection{Visual Feature Representation}\label{sub:visual_repre}
In contrast to previous work \cite{yu-eshghi-lemon:2015:VL,yu-eshghi-lemon:2015:semdial}, to reduce feature noise through the learning process, we simplify the method of feature extraction consisting of two base feature categories, i.e.\ the colour space for colour attributes,  and a `bag of visual words'  for the object shapes/class. 

Colour descriptors, consisting of HSV colour space values, are extracted for each pixel and then are quantized to a $16 \times 4 \times 4$ HSV matrix. These descriptors inside the bounding box are binned into individual histograms. Meanwhile, a bag of visual words is built in PHOW descriptors using a visual dictionary (that is pre-defined with a handmade image set). These visual words will be calculated using 2x2 blocks, a 4-pixel step size, and quantized into 1024 k-means centres. 
The feature extractor in the vision module presents a 1280-dimensional feature vector for a single training/test instance by stacking all quantized features, as shown in Figure \ref{fig:architecture}.

\subsection{Dynamic Syntax and Type Theory with Records}
Dynamic Syntax (DS) a is a word-by-word incremental semantic parser/generator, based around the Dynamic Syntax (DS) grammar framework \cite{Cann.etal05a} especially suited  to the fragmentary and highly contextual nature of dialogue. In DS, dialogue is modelled as the interactive and incremental construction of contextual and semantic representations \cite{Eshghi.etal15}. The contextual representations afforded by DS are of the fine-grained semantic content that is jointly negotiated/agreed upon by the interlocutors, as a result of processing questions and answers, clarification requests, corrections, acceptances, etc. We cannot go into any further detail due to lack of space, but proceed to introduce Type Theory with Records, the formalism in which the DS contextual/semantic representations are couched, but also that within which perception is modelled here.

\paragraph{Type Theory with Records (TTR)} is an extension of standard type theory shown to be useful in semantics and dialogue modelling \cite{Cooper05,Ginzburg12}.
TTR is particularly well-suited to our problem here as it allows information from various modalities, including vision and language, to be represented within a single semantic framework (see e.g.\ Larsson \shortcite{Larsson13}; Dobnik et al.\ \shortcite{Dobnik.etal12} who use it to model the semantics of spatial language and perceptual classification). 
 % as well as its interface with input with various modalities. %\textbf{(Staffan I forget which now)}.

In TTR, logical forms are specified as \emph{record types} (RTs), which are  sequences of \emph{fields} of the form $\smttrnode{}{l&T}$ containing a label $l$ and a type $T$. RTs can be witnessed (i.e.\ judged true) by \emph{records} of that
type, where a record is a sequence of label-value pairs $\smttrrec{}{l&v}$. We say that $\smttrrec{}{l&v}$ is of type $\smttrnode{}{l&T}$ just in case $v$ is of type $T$.

\begin{figure}[!ht]
 \centerline{
 \begin{tabular}{ccc}
 $R_1:\ttrnode{}{l_1 & T_1 \\ l_{2=a} & T_2 \\ l_{3=p(l_2)} & T_3}$
 &
 $R_2:\ttrnode{}{l_1 & T_1 \\ l_2 & T_{2'}}$
 &
 $R_3:\ttrnode{}{}$
 \end{tabular}
 }
 \caption{Example TTR record types}
 \label{fig:ttr}
 \vspace{-.2cm}
 \end{figure}

Fields can be \emph{manifest}, i.e.~given a singleton type e.g.~$\smttrnode{}{l & T_{a}}$ where $T_a$ is the type of which only $a$ is a member; here, we write this using the syntactic sugar $\smttrnode{}{l_{=a}&T}$. Fields can also
be \emph{dependent} on fields preceding them (i.e.~higher) in the record type (see Fig.~\ref{fig:ttr}). 

The standard subtype relation $\subtype$ can be defined for record types: $R_1 \subtype
R_2$ if for all fields $\smttrnode{}{l & T_2}$ in $R_2$, $R_1$ contains $\smttrnode{}{l & T_1}$ where $T_1 \subtype T_2$. In Figure~\ref{fig:ttr}, $R_1 \subtype R_2$ if $T_2 \subtype T_{2'}$, and both $R_1$ and $R_2$ are subtypes of $R_3$. This subtyping relation allows semantic information to be incrementally specified, i.e.\ record types can be indefinitely extended with more information/constraints. For us here, this is a key feature since it allows the system to encode \emph{partial} knowledge about objects, and for this knowledge (e.g.\ object attributes) to be extended in a principled way, as and when this information becomes available.  

\ignore{% not needed for SEMDIAL 2016 paper
\subsection{Dynamic Syntax (DS)} 
The DS module is a word-by-word incremental semantic parser/generator, based around the Dynamic Syntax (DS) grammar framework \cite{Cann.etal05a} especially suited  to the fragmentary and highly contextual nature of dialogue. In DS, dialogue is modelled as the interactive and incremental construction of contextual and semantic representations \cite{Purver.etal11}. The contextual representations afforded by DS are of the fine-grained semantic content that is jointly negotiated/agreed upon by the interlocutors, as a result of processing questions and answers, clarification requests, corrections, acceptances, etc (see Eshghi et al \shortcite{Eshghi.etal15} for an account of how this can be achieved grammar-internally as a low-level semantic update process). Recent versions of DS incorporate Type Theory with Records (TTR) as the logical formalism in which meaning representations are couched \cite{Purver.etal11,Eshghi.etal12}, due to its useful properties. Here we do not introduce DS due to space limitations but proceed to introducing TTR.

%For example, the first row of Fig.~\ref{fig:example} shows the \emph{final} semantic content arrived at jointly by the interlocutors, representing the meaning of: ``Object in joint focus of attention (o1) is red and is a mug". The representations are Record Types (RT) of the Type Theory with Records (TTR) formalism \cite{Cooper05}. TTR is an extension of standard type theory shown useful for modelling the semantics of spatial language, perceptual classification (see e.g.\ \cite{Larsson13}), and dialogue modelling more generally \cite{Ginzburg12}.

\subsection{Type Theory with Records}
\label{sec:ttr}

Type Theory with Records (TTR) is an extension of standard type theory shown to be useful in semantics and dialogue modelling \cite{Cooper05,Ginzburg12}.
TTR is particularly well-suited to our problem here as it allows information from various modalities, including vision and language, to be represented within a single semantic framework (see e.g.\ Larsson \shortcite{Larsson13}; Dobnik et al.\ \shortcite{Dobnik.etal12} who use it to model the semantics of spatial language and perceptual classification). 
 % as well as its interface with input with various modalities. %\textbf{(Staffan I forget which now)}.

In TTR, logical forms are specified as \emph{record types} (RTs), which are  sequences of \emph{fields} of the form $\smttrnode{}{l&T}$ containing a label $l$ and a type $T$. RTs can be witnessed (i.e.\ judged true) by \emph{records} of that
type, where a record is a sequence of label-value pairs $\smttrrec{}{l&v}$. We say that $\smttrrec{}{l&v}$ is of type $\smttrnode{}{l&T}$ just in case $v$ is of type $T$.

\begin{figure}[!ht]
 \centerline{
 \begin{tabular}{ccc}
 $R_1:\ttrnode{}{l_1 & T_1 \\ l_{2=a} & T_2 \\ l_{3=p(l_2)} & T_3}$
 &
 $R_2:\ttrnode{}{l_1 & T_1 \\ l_2 & T_{2'}}$
 &
 $R_3:\ttrnode{}{}$
 \end{tabular}
 }
 \caption{Example TTR record types}
 \label{fig:ttr}
 \vspace{-.2cm}
 \end{figure}

Fields can be \emph{manifest}, i.e.~given a singleton type e.g.~$\smttrnode{}{l & T_{a}}$ where $T_a$ is the type of which only $a$ is a member; here, we write this using the syntactic sugar $\smttrnode{}{l_{=a}&T}$. Fields can also
be \emph{dependent} on fields preceding them (i.e.~higher) in the record type (see Fig.~\ref{fig:ttr}). 

The standard subtype relation $\subtype$ can be defined for record types: $R_1 \subtype
R_2$ if for all fields $\smttrnode{}{l & T_2}$ in $R_2$, $R_1$ contains $\smttrnode{}{l & T_1}$ where $T_1 \subtype T_2$. In Figure~\ref{fig:ttr}, $R_1 \subtype R_2$ if $T_2 \subtype T_{2'}$, and both $R_1$ and $R_2$ are subtypes of $R_3$. This subtyping relation allows semantic information to be incrementally specified, i.e.\ record types can be indefinitely extended with more information/constraints. For us here, this is a key feature since it allows the system to encode \emph{partial} knowledge about objects, and for this knowledge (e.g.\ object attributes) to be extended in a principled way, as and when this information becomes available.  

%Following \citet{Purver.etal11}, we assume that DS tree nodes are decorated with RTs, and corresponding lambda abstracts representing functions from RT to RT (e.g.~$\lambda r\!\!:\!\smttrnode{}{l_1&T_1}.\smttrnode{}{l_{2=r.l_1}&T_1}$ where $r.l_1$ is a \emph{path} expression referring to the label $l_1$ in $r$) -- see Figure~\ref{tree}. %\textbf{CH: I don't understand this sentence, can we lose it?: The equivalent of conjunction for linked trees is now RT \emph{extension}  \citep[concatenation modulo relabelling -- see][]{Cooper05,Fernandez06}.} 
%TTR's subtype relation allows a record type at the root node to be inferred for any partial tree, and incrementally further specified via subtyping as parsing proceeds \citep{Hough.Purver12}.
}  %%%%%% END OF IGNORE

\subsection{Integration}

Fig.~\ref{fig:architecture} shows how the various parts of the system interact. At any point in time, the system has access to an ontology of (object) types and attributes encoded as a set of TTR Record Types, whose individual atomic symbols, such as `red' or `square' are grounded in the set of classifiers trained so far. 

\begin{figure*}[!t]
  \centering
  \includegraphics[width=0.8\linewidth]{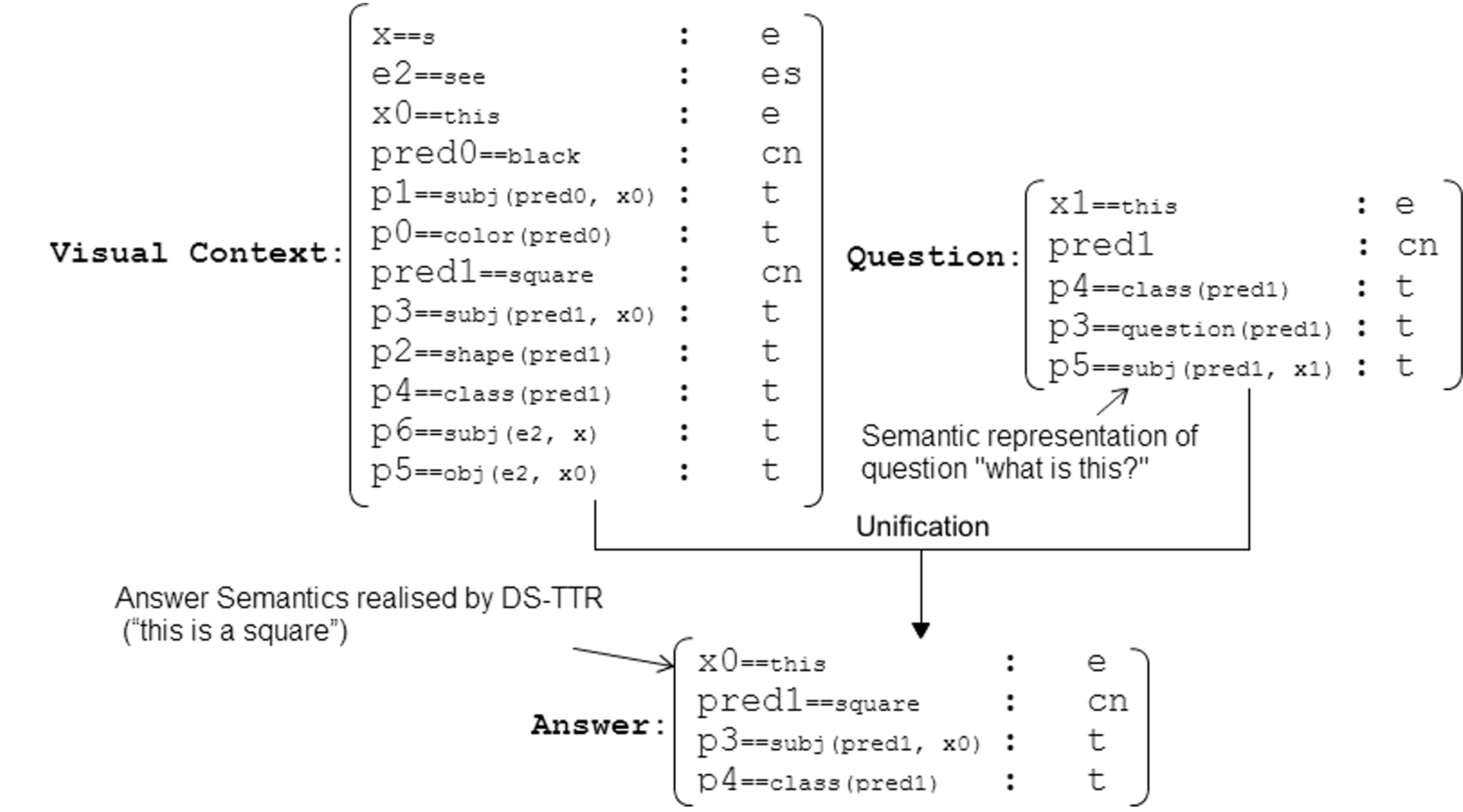}
  \vspace{-0.2cm}
  \caption{Question Answering by the system}\label{fig:qa_example}
\end{figure*}

Given a set of individuated objects in a scene, encoded as a TTR Record, the system can utilise its existing ontology to output a Record Type which maximally characterises the scene (see e.g.\ Fig.~\ref{fig:example}). Dynamic Syntax operates over the same representations, they provide a direct interface between perceptual classification and semantic processing in dialogue: this representation acts not only as (1) the non-linguistic (here, visual) context of the dialogue for the resolution of e.g. definite reference and indexicals (see \cite{Hough.Purver14}); but also as (2) the logical database from which the system can generate utterances (descriptions), ask, or answer questions about the objects - Fig. \ref{fig:qa_example} illustrates how the semantics of the answer to a question is retrieved from the visual context through \emph{unification} (this uses the standard sub-type checking operation within TTR).

Conversely, for concept learning, the DS-TTR parser incrementally produces Record Types (RT), representing the meaning jointly established by the tutor and the system so far. In this domain, this is ultimately one or more type judgements, i.e.\ that some scene/image/object is judged to be of a particular type, e.g.\ in Fig.~\ref{fig:example} that the individuated object, $o1$ is a red square. These jointly negotiated type judgements then go on to provide training instances for the classifiers. In general, the training instances are of the form, $\langle O, T\rangle$, where $O$ is an image/scene segment (an object or TTR Record), and $T$, a record type. $T$ is then decomposed into its constituent atomic types $T_1\ldots T_n$, s.t. $\bigwedge T_i=T$ - where $\bigwedge$ is the so called \emph{meet} operation corresponding to type conjunction. The judgements $O:T_i$ are then used directly to train the classifiers that ground the $T_i$. %e.g.\ the dialogues in Fig.~\ref{fig:example} provide the following instances to our classifiers: $\langle o1, \{red, mug\}\rangle$ and $\langle o2, \{red, book\}\rangle$.

%% YYC: I have included all possible dialogue (we mentioned in this paper) into this example. Please let me know if there is anything new you want to add into...

\begin{figure*}[ht]
\centering

\includegraphics[width=\linewidth]{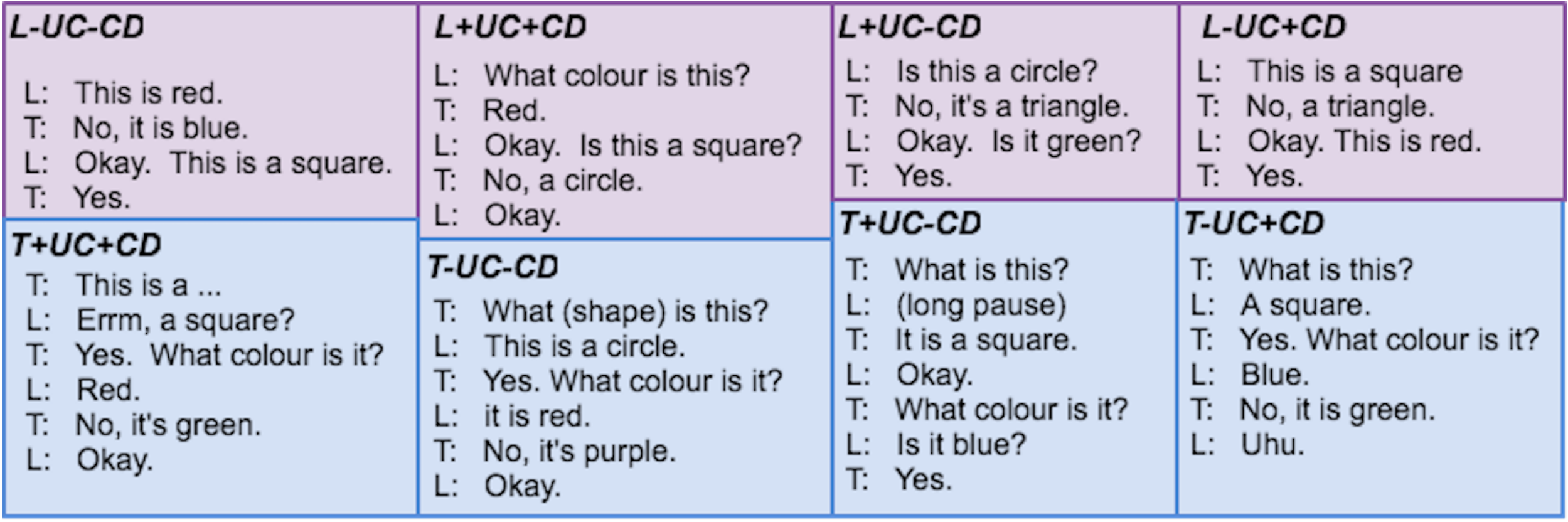}
\caption{Example dialogues in different conditions}\label{fig:example_dialogue}
\end{figure*}
\section{\label{sec:expriment_setup} Experimental Setup}

% . L/T= learner/tutor initiative; UC= uncertainty processing; CD= context-dependency 

As noted in the introduction, interactive systems that learn continuously, and over the long run from humans need to do so \textit{incrementally}; \textit{as quickly as possible}; and \textit{with as little effort/cost to the human tutor as possible}. In addition, when learning takes place through dialogue, the dialogue needs to be as human-like/natural as possible.

In general, there are several different dialogue capabilities and policies that a concept-learning agent might adopt, and these will lead to different outcomes for the accuracy of the learned concepts/meanings, learning rates, and cost to the tutor --  with trade-offs between these. Our goal in this paper is therefore  an experimental study of the effect of different dialogue policies and capabilities on the overall performance of the learning agent, which, as we describe below is a measure  capturing the trade-off between accuracy of learned meanings and the cost of tutoring.
% over time.

\paragraph{Design.} We use the dialogue system outlined above to carry out our main experiment with a $2 \times 2 \times 2$ factorial design, i.e.\ with three factors each with two levels. Together, these factors determine the learner's dialogue behaviour: (1) \textbf{Initiative} (\textbf{L}earner/\textbf{T}utor): determines who takes initiative in the dialogues. When the tutor takes initiative, s/he is the one that drives the conversation forward, by asking questions  to the learner (e.g.\ ``What colour is this?'' or ``So this is a ....'' ) or making a statement about the attributes of the object. On the other hand, when the learner has initiative, it  makes statements, asks questions, initiates topics etc. (2) \textbf{Uncertainty} (\textbf{+UC}/\textbf{-UC}): determines whether the learner takes into account, in its dialogue behaviour, its own subjective confidence about the attributes of the presented object. The confidence is the probability assigned by any of its  attribute classifiers of the object being a positive instance of an attribute (e.g.\ `red') - see below for how a confidence threshold is used here. In +UC, the agent will not ask a question if it is confident about the answer, and it will hedge the answer to a tutor question if it is not confident, e.g.\ ``T: What is this? L: errm, maybe a square?". In -UC, the agent always takes itself to know the attributes of the given object (as given by its currently trained classifiers), and behaves according  to that assumption. (3) \textbf{Context-Dependency} (\textbf{+CD}/\textbf{-CD}): determines whether the learner can process (produce/parse) context-dependent expressions such as short answers and incrementally constructed turns, e.g.\ ``T: What is this? L: a square", or ``T: So this one is ...? L: red/a circle''. This setting  can be turned off/on in the DS-TTR dialogue model.

\paragraph{Tutor Simulation and Policy} To run our experiment on a large-scale, we have hand-crafted an \textit{Interactive Tutoring Simulator}, which simulates the behaviour of a human tutor\footnote{The experiment involves hundreds of dialogues, so running this experiment with real human tutors has proven too costly at this juncture, though we plan to do this for a full evaluation of our system in the future.}. The tutor policy is kept constant across all conditions. Its policy is that of an always \textit{truthful}, \textit{helpful} and \textit{omniscient} one: it (1) has complete access to the labels of each object; and (2) always acts as the context of the dialogue dictates: answers any question asked, confirms or rejects when the learner describes an object; and (3) always corrects the learner when it describes an object erroneously.

\paragraph{Dependent Measures} We now go on to describe the dependent measures in our experiment, i.e.\ that of classifier accuracy/score, tutoring cost, and the overall performance measure which combines the former two measures.

%% YYC: confidence threshold consists of two values: positive threshold and negative threshold (0.5). I just describe them separately below.  

\paragraph{Confidence Threshold} To determine when the agent takes themselves to be confident in an attribute prediction, we use confidence-score thresholds.
%involves into a cooperative behaviour with the \textit{Uncertainty} factor. 
It consists of two values, a base threshold (e.g.\ 0.5)  and a positive threshold (e.g.\ 0.9). 
%The former value (as 0.5 in the experiment) is to distinguish positive and negative predictions -- 

If the confidences of all classifiers are under the base threshold (i.e.\ the learner has no  attribute label that it is confident about), the agent will ask for  information directly from the tutor via questions (e.g.\ ``L: what is this?").
% instead of waywardly picking up one prediction with the highest confidence, e.g. "T: what is this? L: Sorry, I don't know it. Can you tell me? ". 

On the other hand, if one or more classifiers score above the base threshold, then 
%as described with the \textit{Uncertainty},
 the positive threshold is used to judge to what extent the agent trusts its prediction or not. If the confidence score of a classifier is between the positive and base thresholds, the  learner is not very confident about its knowledge, and will check with the tutor, e.g.\ ``L: is this red?''. However, if the   confidence score of a classifier is above the positive threshold, the  learner is confident enough in its knowledge not to bother verifying it with the tutor. This will lead to less effort needed from the tutor as the learner becomes more confident about its knowledge. However, since a learning agent that has high confidence about a prediction will not ask for  assistance from the tutor, a low positive threshold would reduce the chances that allow the tutor to correct the learner's mistakes. We therefore tested different fixed values for the confidence threshold and this determined a fixed 0.5 base threshold and a 0.9 positive threshold were deemed to be the most appropriate values for an interactive learning process - i.e.\ these values preserved good classifier accuracy while not requiring much effort from the tutor - see below Section \ref{sec:adaptive} for how an adaptive policy was learned that adjusts the agent's confidence threshold dynamically over time.

% from the agent's predictions. 
%Hence, we set up an auxiliary experiment, in which we kept all other conditions constant (i.e.\ assume that the learner has initiative (\textbf{L}) and always considers the prediction confidence(\textbf{+U})), but only varied the threshold values.
% (e.g. 0.7, 0.8, 0.9 and 0.95).
%This additional experiment determined a 0.5 base threshold and a 0.9 positive threshold as the most appropriate values for an interactive learning process - i.e.\ this preserved good classifier accuracy while not requiring much effort from the tutor.
%OL - I  wonder should we show the graph of this??? 

\subsection{Evaluation Metrics} 

To test how the different dialogue capabilities and strategies   affect the   learning process, we consider both the cost to the tutor and the accuracy of the learned meanings, i.e.\ the classifiers that ground our colour and shape concepts.

% AE: commented out, we are no longer really following Skicaj etal.
%we   follow metrics proposed by Skocaj et al.\ \shortcite{Skocaj2009a}, that consist of two main evaluation measures, i.e.\ {\it Recognition Scores}  and {\it Tutoring Costs}. We modify the details below to accommodate our  different dialogue system configurations.

%AE:commented out. We are no longer using Recognition Score, but the standard Accuracy measure.
%  \subsubsection{Recognition score}
% This is a metric measuring the overall accuracy of the learned word meanings / classifiers, which ``rewards successful classifications (i.e.\ true positives and true negatives) and penalizes incorrect predictions (i.e.\ false positives and false negatives)" \cite{Skocaj2009a}. As the proposed system considers both correctness of predicted labels and prediction confidence on learning tasks, the measure will also take the true labels with lower confidence into account, as shown in Table \ref{tab:recogScore}; ``LowYes'' means that the system made positive predictions but with lower confidence. In this case, the system can generate a polar question for requesting tutor feedback. ``LowNo'' is similar to ``LowYes'', but only works on negative predictions.

% \begin{table}[!t]
% \centering
% \caption{\label{tab:recogScore} Recognition Score Table}
% \begin{tabular}[width=\columnwidth]{ccccc}
% 	\hline
%      & Yes & LowYes & LowNo & No\\
% 	\hline
%     Yes & 1 & 0.5 & -0.5 & -1 \\
%     No & -1 & -0.5 & 0.5 & 1 \\
%     \hline
% \end{tabular}
% \end{table}
 
\paragraph{Cost} The cost measure reflects the effort needed by a human tutor in interacting with the system. Skocaj et.\ al.\ \shortcite{Skocaj2009a} point out that a comprehensive teachable system should learn as autonomously as possible,  rather than involving the human tutor too frequently. There are several possible costs that the tutor might incur, see  Table \ref{tab:tutorcost}: $C_{inf}$ refers to the cost of the tutor providing information on a single attribute concept (e.g.\ ``this is red'' or ``this is a square"); $C_{ack}$ is the cost for a simple confirmation (like ``yes", ``right") or rejection (such as ``no''); $C_{crt}$ is the cost of correction for a single concept (e.g.\ ``no, it is blue" or ``no, it is a circle"). We associate a higher cost with correction of statements than that of polar questions. This is to penalise the learning agent when it confidently makes a false statement --  thereby incorporating an aspect of trust in the metric (humans will not trust systems which confidently make false statements). And finally, parsing ($C_{parse}$) as well as production ($C_{production}$) costs for tutor are taken into account: each single word costs 0.5 when parsed by the tutor, and 1 if generated (production costs twice as much as parsing). These exact values are based on intuition but are kept constant across the experimental conditions and therefore do not confound the results reported below.

% that it is just as much effort for the Tutor to provide a concept as to correct one, and that confirmation has a smaller cost, while each turn also requires a small effort from the Tutor. The experiment considers a cumulative tutoring cost through the entire learning process with 500 training instances. The costs are equal across all the conditions and therefore do not confound the results reported below.
%These settings implement the intuition that  The results below still hold as long as these relative costs are maintained.  

%Finally, the tutor will not experience any cost while ignoring learners' statements or requests. 

% We define the overall tutoring cost at particular learning steps as:
% \begin{displaymath}
% {\scriptstyle C_{tutor} = \sum_{n=1}C_{inf} + \sum_{n=1}C_{yes} +\sum_{n=1}C_{crt} + \sum_{n=1}C_{parse} + \sum_{n=1}C_{pruduct}}
% \end{displaymath}

\begin{table}[!t]
\centering
\caption{\label{tab:tutorcost} Tutoring Cost Table}
\begin{tabular}[width=0.8\columnwidth]{ccccc}
	\hline
    $C_{inf}$ & $C_{ack}$ & $C_{crt}$ & $C_{parsing}$ & $C_{production}$\\
	\hline
    1 & 0.25 & 1 & 0.5 & 1 \\
    \hline
\end{tabular}
\end{table}

% \begin{figure*}[!ht]
% \subfloat[Recognition Score\label{fig:gt_recgScore}]
%   {\includegraphics[width=.5\linewidth]{exptResults/perform_gt.jpg}}\hfill
% \subfloat[Tutoring Cost\label{fig:gt_cost}]
%   {\includegraphics[width=.5\linewidth]{exptResults/cost_gt.jpg}}\hfill
% %\subfloat[Performance/Cost Ratio\label{fig:gt_fianl}]
%  % {\includegraphics[width=.33\linewidth]{exptResults/finalperform_gt.jpg}}
% \caption{Evolution of Learning Performance in the \textit{Good Tutor} Condition}\label{fig:results_gt}
% \end{figure*}

\paragraph{Learning Performance} As mentioned above, an efficient learner dialogue policy should consider both classification accuracy  and tutor effort (Cost). We thus define an integrated measure -- the \textit{Overall Performance Ratio} ($R_{perf}$) -- that we use to compare the learner's overall performance across the different conditions:

\vspace{-0.4cm}
\begin{displaymath}
 R_{perf} = \frac{\Delta Acc}{C_{tutor}}
\end{displaymath}

\noindent i.e.\ the increase in accuracy per unit of the cost, or equivalently the gradient of the curve in Fig.\ 4c.\ We seek dialogue strategies that maximise this.

%% YYC: shall we briefly introduce the handmade dataset we used in this experiment??

\paragraph{Dataset} The dataset used here is comprised of 600 images of single, simple handmade objects with a white background (see Fig.\ref{fig:example})\footnote{All data from  this paper will be made freely available.}. There are nine attributes considered in this dataset: 6 colours (black, blue, green, orange, purple and red) and 3 shapes (circle, square and triangle), with a relative balance on the number of instances per attribute.

%AE: shortened the below

%In order to compare the different dialogue capabilities and strategies, we built up a dataset of 600 images of simple handmade objects\footnote{All data from  this paper will be made freely available.}. The goal of this system was to learn simple visual attributes (e.g.\ colour and shape) from these simple objects (see two object examples in Fig.\ref{fig:example}). There are nine attributes considered in this dataset: 6 colours (black, blue, green, orange, purple and red) and 3 shapes (circle, square and triangle). As background noise might interfere with the ability of object segmentation and extraction, we build images containing only one object within a white background. The system can then automatically detect object boundaries and build the corresponding perceptual representations as described in section \ref{sub:visual_repre}. We keep a relative balance on the number of instances for each attribute in the dataset.  
     
% \begin{figure}[!t]
% 	\centering
% 	\includegraphics[width=0.35\textwidth]{imageResource/ObjExamples.png}
% 	\vspace{-0.2cm}
%     \caption{Examples of simple handmade objects}\label{fig:objExample}
% \end{figure}

% To ensure that each attribute-based classifier can be learned with enough positive examples, we kept a relative balance between different attributes in the same categories, i.e.\ 100 positive instances of each colour property, 147 instances of ``circle", 232 instances of ``square", and 221 instances of ``triangle". 

\subsection{Evaluation and Cross-validation}
In each round, the system is trained using 500 training instances, with the rest set aside for testing. For each training instance, the system interacts (only through dialogue) with the simulated tutor. Each dialogue about an object ends either when both the shape and the colour of the object are discussed and agreed upon, or when the learner requests to be presented with the next image (this happens only in the Learner initiative conditions). We define a \textbf{Learning Step} as comprised of 10 such dialogues. At the end of each learning step, the system is tested using the test set (100 test instances). 

This process is repeated 20 times, i.e.\ for 20 rounds/folds, each time with a different, random 500-100 split, thus resulting in 20 data-points for cost and accuracy after every learning step. The values reported below, including those on the plots in Fig.\ 6a, 6b and 6c, correspond to averages across the 20 folds.

%OL - be careful about these figure references! are they correct?

 \begin{figure*}[!ht]
\subfloat[Accuracy\label{fig:correct_recgScore}]
  {\includegraphics[width=.5\linewidth]{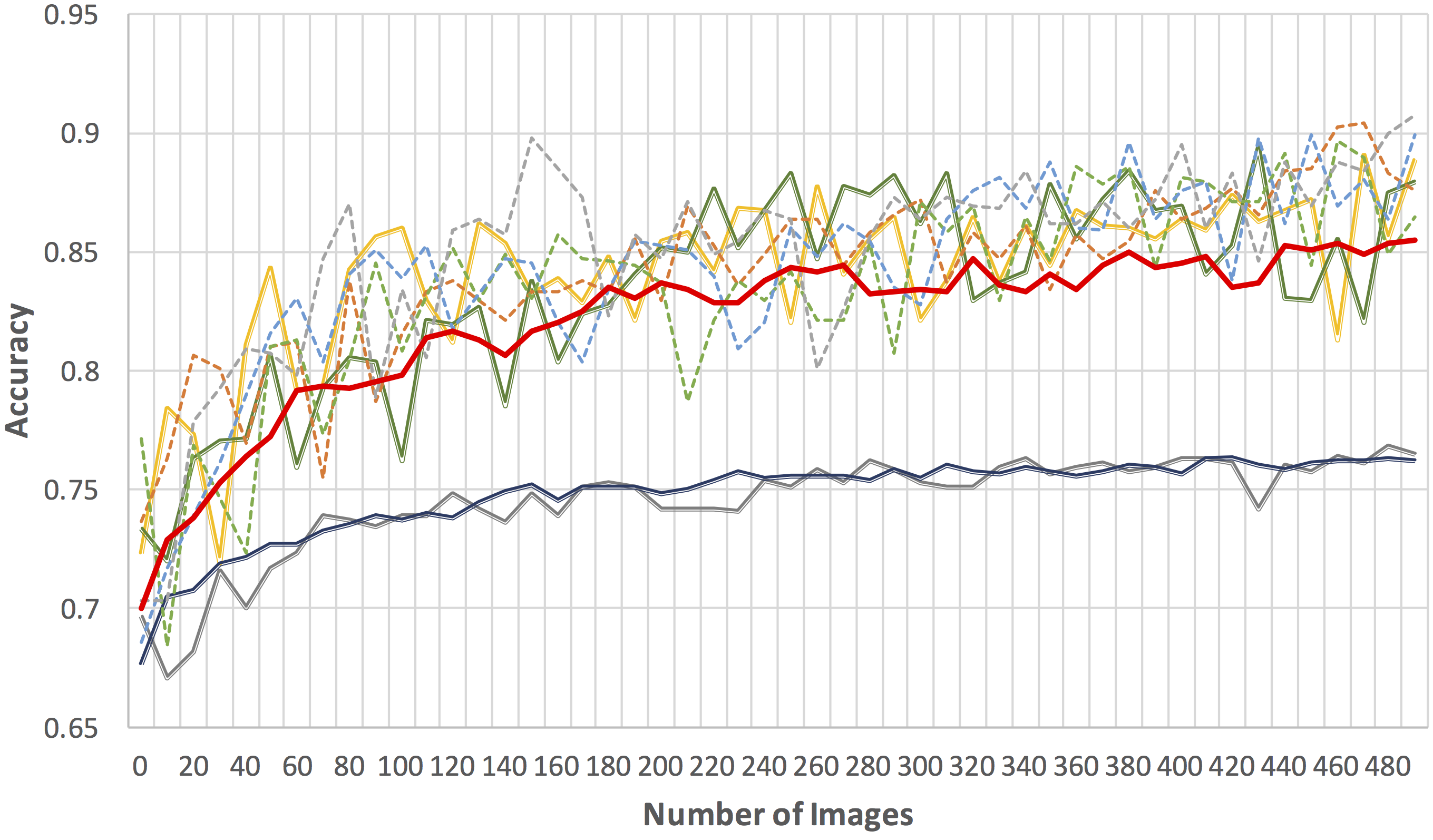}}\hfill
  \subfloat[Tutoring Cost\label{fig:correct_cost}]
  {\includegraphics[width=.5\linewidth]{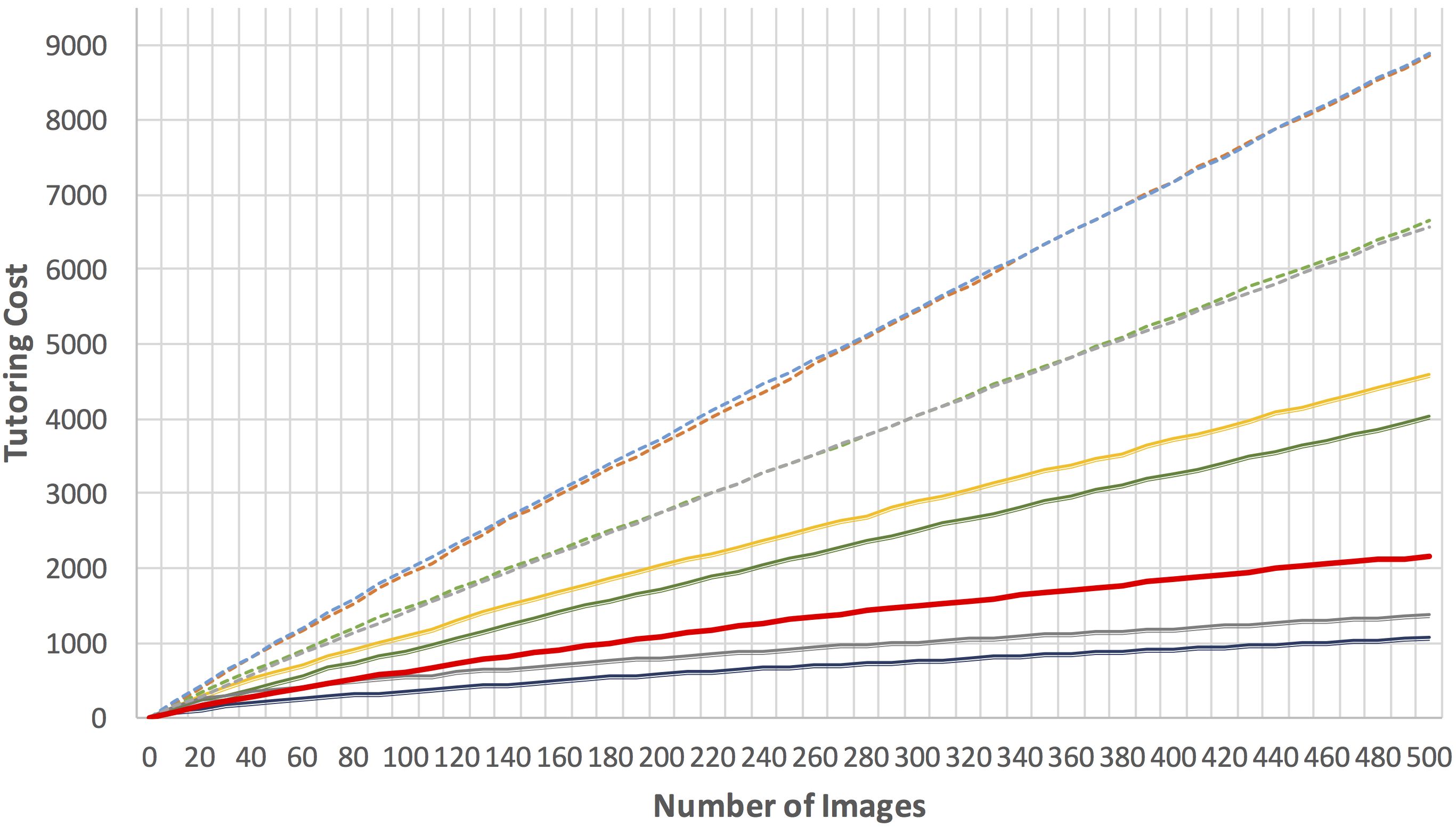}}\hfill
  
\centering  
\subfloat[Overall Performance\label{fig:correct_fianl}]
 {\includegraphics[width=.75\linewidth]{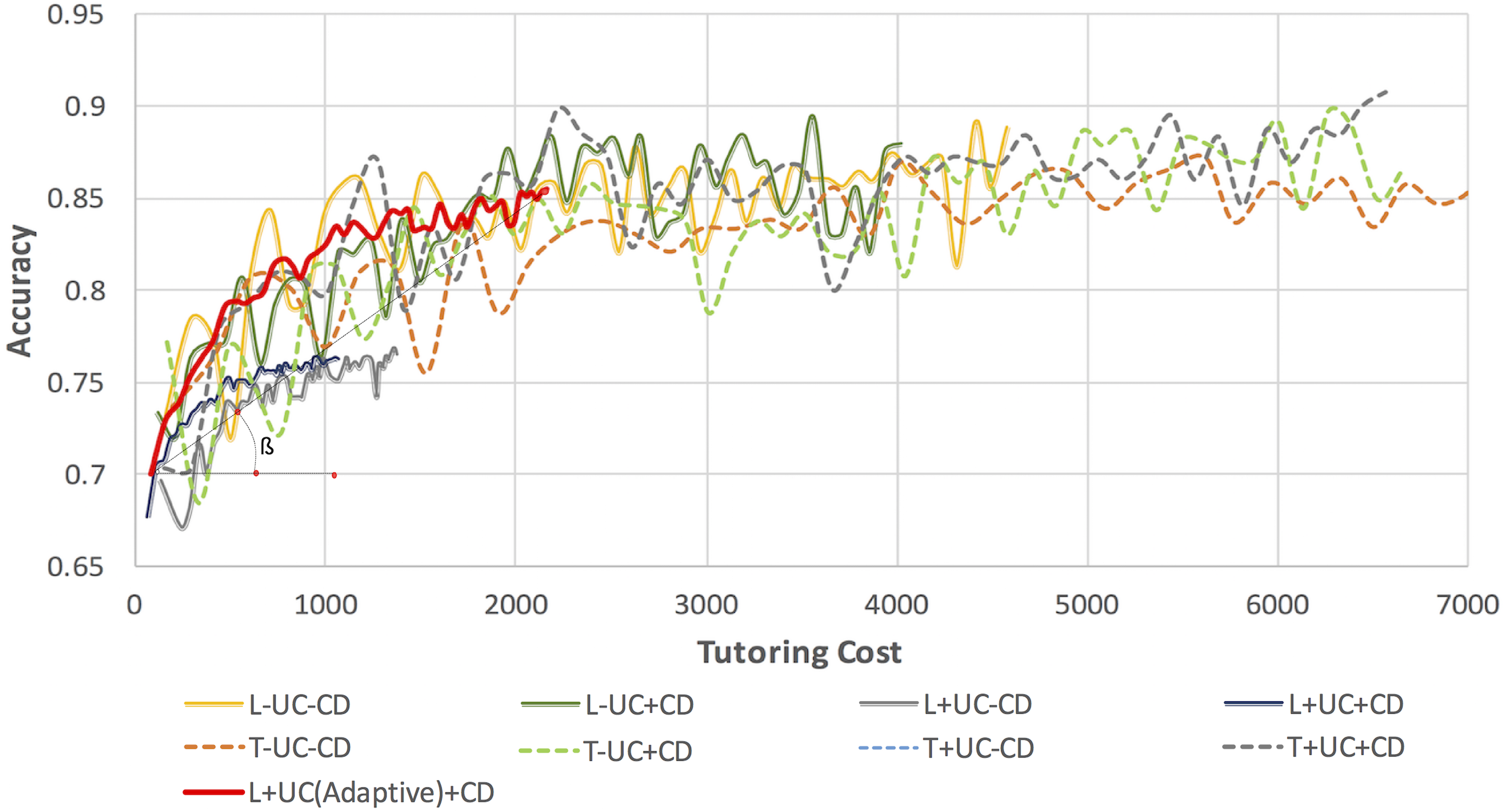}}
\caption{Evolution of Learning Performance}
%in the \textit{Correction} Condition} %OL removed
\label{fig:results_lt}
\end{figure*}

\subsection{Learning an Adaptive Policy for a Dynamic Confidence Threshold\label{sec:adaptive}} 
%AE: Yanchao, please edit this section as appropriate.
In the   experiment  presented above, the learning agent's positive confidence threshold was held constant, at  0.9. However, since the confidence threshold itself becomes more reliable as the agent is exposed to more training instances, we further hypothesised that a threshold that changes dynamically over time should lead to a better trade-off between classification accuracy and cost for the tutor, i.e.\ a better \textit{Overall Performance Ratio} (see above). For example, lower positive thresholds may be more appropriate at the later stages of training when the agent is already performing well with attribute classifiers which are more reliable. This leads to different dialogue behaviours, as the learner takes different decisions as it encounters more training examples.

To test this hypothesis 
%that a dynamic threshold leads to a better trade-off between accuracy and cost, 
we further trained and evaluated an adaptive policy that adjusts the learning agent's confidence threshold as it interacts with the tutor (in the +UC conditions only). This optimization used a Markov Decision Process (MDP) model and Reinforcement Learning\footnote{A reviewer points out that one can handle uncertainty in a more principled way, possibly with better results, using POMDPs. Another reviewer points out that the policy learned is only adapting the confidence threshold, and not the other conditions (uncertainty, initiative, context-dependency). We point out that we are addressing both of these limitations in work in progress, where we feed each classifier's outputted confidence level as a continuous feature in a (continuous space) MDP for \emph{full dialogue control}.}, where: (1) the \textbf{state space} was determined by variables for the number of training instances seen so far, and the agent's current confidence threshold (2) the \textbf{actions} were  either to increase or decrease the confidence threshold by 0.05, or keep it the same; (3) the \textbf{local reward signal} was directly proportional to the agent's \textit{Overall Performance Ratio} over the previous Learning Step (10 training instances, see above); and (4) the {\bf SARSA algorithm} \cite{Sutton.Barto98} was chosen for learning, with each \textbf{episode} defined as a complete run through the 500 training instances.

% Note: Through this auxiliary experiment, we kept all policy conditions constantly (i.e. the Learner takes initiative with both uncertainty and context-dependency switched on (L+UC+CD)), but with different threshold policy, i.e. Constant or Adaptive.  

%AE: whom should we cite for SARSA? OL: Sutton and Barto RL book

\section{\label{sec:results}Results} 
Fig.\ \ref{fig:example_dialogue} shows example interactions between the learner and the tutor in some of the experimental conditions. Note how the system is able to deal with (parse and generate) utterance continuations as in $T+UC+CD$, short answers as in $L+UC+CD$, and polar answers as in $T+UC+CD$.  

Fig.\ 6a and 6b plot the progression of average Accuracy and (cumulative) Tutoring Cost for each of the 8 conditions in our main experiment, as the system interacts over time with the tutor about each of the 500 training instances. The ninth curve in red (L+UC(Adaptive)+CD) shows the same for the learning agent with a dynamic confidence threshold using the policy trained using Reinforcement Learning (section \ref{sec:adaptive}) - the latter is only compared below to the dark blue curve (L+UC+CD). As noted in passing, the vertical axes in these graphs are based on averages across the 20 folds - recall that for Accuracy the system was tested, in each fold, at every learning step, i.e.\ after every 10 training instances.

Fig.\ 6c, on the other hand, plots Accuracy against Tutoring Cost directly. Note that it is to be expected that the curves should not terminate in the same place on the x-axis since the different conditions incur different total costs for the tutor across the 500 training instances. The gradient of this curve corresponds to \textit{increase in Accuracy per unit of the Tutoring Cost}. It is the gradient of the line drawn from the beginning to the end of each curve ($tan(\beta)$ on Fig.\ 4c) that constitutes  our main evaluation measure of the system's overall performance in each condition, and it is this measure for which we report statistical significance results:

A between-subjects Analysis of Variance (ANOVA) shows significant main effects of Initiative ($p<0.01; F = 448.33$), Uncertainty ($p<0.01; F = 206.06$) and Context-Dependency ($p<0.05; F = 4.31$) on the system's overall performance. There is also a significant Initiative$\times$Uncertainty interaction ($p<0.01; F = 194.31$).

Keeping all other conditions constant (L+UC+CD), there is also a significant main effect of Confidence Threshold type (Constant vs. Adaptive) on the same measure ($p<0.01; F = 206.06$). The mean gradient of the red, adaptive curve is actually slightly lower than its constant-threshold counter-part blue curve - discussed below. 

\section{\label{sec:discussion}Discussion}
\paragraph{Tutoring Cost} As can be seen on Fig.\ 6b, the cumulative cost for the tutor progresses more slowly when the learner has initiative (L) and takes  its confidence into account in its behaviour (+UC) - the grey, blue, and red curves. This is so because \textit{a form of active learning} is taking place: the learner only asks a question about an attribute if it isn't confident enough already about that attribute. This also explains the slight decrease in the gradients of the  curves as the agent is exposed to more and more training instances: its subjective confidence about its own predictions  increases over time, and thus there is progressively less need for tutoring.

\paragraph{Accuracy} On the other hand, the L+UC curves (grey and blue) on Fig.\ 6a show the slowest increase in accuracy and flatten out at about 0.76. This is because the agent's confidence score in the beginning is unreliable as the agent has only seen a few training instances: in many cases it doesn't query the tutor or have any interaction whatsoever with it and so there are informative examples that it doesn't get exposed to. In contrast to this, the L+UC(adaptive)+CD curve (red) achieves much better accuracy. 

Comparing the gradients of the curves on Fig.\ 6c shows that the overall performance of the agent on the gradient measure is significantly better than others in the L+UC conditions (recall the significant Initiative $\times$ Uncertainty interaction). However, while the agent with an {\bf adaptive} threshold (red/L+UC(adaptive)+CD) achieves slightly lower overall gradient than its constant threshold counter-part (blue/L+UC+CD), it achieves much higher Accuracy overall, and does this much faster in the first 1000 units of   cost (roughly the total cost in L+UC+CD condition). We therefore conclude that the adaptive policy is more desirable.
% \paragraph{Context-Dependency} 
Finally, the significant main effect of {\bf Context-Dependency} on the overall performance is explained by the fact in the +CD conditions, the agent is able to process context-dependent and incrementally constructed turns, leading to less repetition, shorter dialogues, and therefore better overall performance.

\section{Conclusion and Future work}

We have presented a multi-modal dialogue system that learns grounded word meanings from a human tutor, incrementally, over time, and employs a dynamic  dialogue policy (optimised using Reinforcement Learning). The system integrates a semantic grammar for dialogue (DS), and a logical theory of types (TTR), with a set of visual classifiers in which the TTR semantic representations are grounded. We used this implemented system to study the effect of different dialogue policies and capabilities on the overall performance of a learning agent - a combined measure of accuracy and cost. The results show that in order to maximise its performance, the agent needs to take initiative in the dialogues, take into account its changing confidence about its predictions, and be able to process natural, human-like dialogue.

%We used static policies in our experiments here. 
Ongoing work further uses Reinforcement Learning to learn complete, incremental dialogue policies, i.e.\ which choose system output at the lexical level \cite{Eshghi.Lemon.2014}. To deal with uncertainty this system takes all the classifiers' outputted confidence levels directly as features in a continuous space MDP.
% that optimise such an agent's performance.
%,  and comparing the learned policies with those of this study.

%\section*{Acknowledgements} for final version 
 
%\section*{Acknowledgements} for final version 
 %\vspace{-0.25cm}
\section*{Acknowledgements}
 This research is  supported by the EPSRC, under grant number EP/M01553X/1 (BABBLE project\footnote{\url{https://sites.google.com/site/hwinteractionlab/babble}}),
and by the European Union's Horizon 2020 research and innovation programme under grant agreement No.\ 688147 (MuMMER project\footnote{\url{http://mummer-project.eu/}}).

\bibliographystyle{acl}
\bibliography{all,SG_B,babble,thesis_all}

 \end{document}